\documentclass{article}
\usepackage{spconf,amsmath,graphicx,multicol}
\usepackage{amsfonts}
\usepackage{bbm}
\usepackage{cite}
\usepackage{amssymb}
\usepackage{algorithm}
\usepackage{algorithmic}
\usepackage{tabularx}
\usepackage[utf8]{inputenc}
\usepackage[english]{babel}
\usepackage{amsthm}
\usepackage{subcaption}
\theoremstyle{definition}

\newtheorem{theorem}{Theorem}[section]

\let\oldemptyset\emptyset
\let\emptyset\varnothing
\newcommand\norm[1]{\left\lVert#1\right\rVert}
\newcommand{\argmax}{\arg\!\max}
\newcommand{\argmin}{\arg\!\min}

\def\x{{\mathbf x}}

\def\y{{\mathbf y}}
\def\e{{\mathbf e}}

\def\A{{\mathbf A}}
\def\B{{\mathbf B}}
\def\P{{\mathbf P}}

\def\R{{\mathbb{R}}}

\def\I{{\mathbf I}}
\def\a{{\mathbf a}}

\usepackage{mathtools}

\title{Sparse Linear Regression Via Generalized Orthogonal Least-Squares}
\name{Abolfazl Hashemi and Haris Vikalo}
\address{Department of Electrical and Computer Engineering\\
		University of Texas at Austin, Austin, TX, USA}
\begin{document}
%
\maketitle
\begin{abstract}
The Orthogonal Least Squares (OLS) algorithm sequentially selects columns of the coefficient 
matrix to greedily find an approximate sparse solution to an underdetermined system of linear equations. 
In this paper, conditions under which OLS recovers sparse signals from a low number of random linear 
measurements with probability arbitrarily close to one are stated. Moreover, a computationally efficient 
generalization of Orthogonal Least-Squares which relies on a recursive relation between the components 
of the optimal solution to select $L$ columns at each step and solve the resulting overdetermined 
system of equations is proposed. This generalized OLS algorithm is empirically shown to outperform 
existing greedy algorithms broadly used in literature. 
\end{abstract}
\section{Introduction}
\label{sec:intro}
The problem of finding sparse solution to an underdetermined system of linear equations arises in a number 
of practical scenarios. Examples include compressed sensing \cite{donoho2006compressed}, sparse channel
estimation in communication systems \cite{mitr07,barik2014sparsity}, compressive DNA microarrays 
\cite{parvaresh2008recovering} as well as a number of other applications in signal processing and machine 
learning \cite{lustig2007sparse,elad2010role,mishali2010theory,elhamifar2009sparse}. Consider the linear 
measurement model
\begin{equation} \label{eq:1}
{\y=\A\x+{\bf e}},
\end{equation}
where $\y \in\R^{n}$ denotes the vector of observations, $\A \in\R^{n \times m}$ is the coefficient matrix (i.e., a collection of features) assumed to be full rank, ${\bf e} \in\R^{n}$ is the additive observation noise vector, and $\x \in\R^{m}$ is a vector known to have at most $k$ non-zero components (i.e., $k$ is the sparsity level of $\x$).  We are interested in finding a sparse approximation to $\x$; in particular, we would like to solve the so-called $l_0$-constrained least-squares
\begin{equation}  \label{eq:2}
\begin{aligned}
& \underset{\x}{\text{minimize}}
&  \norm{\y-\A\x}^{2}_{2}
& &\text{subject to}
& & \norm{\x}_{0} \leq k.
\end{aligned}
\end{equation}
The number of possible locations of non-zero entries in $\x$ scales combinatorially with $n$ which renders
(\ref{eq:2}) computationally challenging; in fact, the problem is NP-hard. To enable computationally efficient 
search for sparse $\x$ approximating (\ref{eq:1}), the non-convex $l_0$-norm-constrained optimization
(\ref{eq:2}) can be replaced by a sparsity-promoting $l_1$-norm optimization
\begin{equation} 
\begin{aligned}
& \underset{\x}{\text{minimize}}
& \norm{\x}_{1}
& & \text{subject to}
& & \norm{\y - \A \x}_2 \leq \varepsilon,
\end{aligned}
\label{eq:L1}
\end{equation}
where $\varepsilon > 0$ is a predetermined measure of noise power. In the noise-free scenario where
$\e$ in (\ref{eq:1}) and $\varepsilon$ in (\ref{eq:L1}) are both zero and where $\A$ satisfies certain properties, 
it is known that a sufficiently sparse $\x$ can be reconstructed exactly \cite{candes2006robust}. 
However, while the convexity of $l_{1}$-norm enables finding 
the optimal solution to the reformulated sparse vector recovery problem, the complexity of doing so (by 
means of, e.g., iterative shrinkage-thresholding algorithms such as \cite{beck2009fast}, or alternating 
direction method of multipliers \cite{boyd2011distributed}) is often prohibitive when one
deals with high dimensional data. For this reason, a number of fast greedy heuristics that attempt to solve 
(\ref{eq:2}) directly by successively identifying columns of $\A$ which correspond to non-zero components 
of $\x$ have been proposed \cite{tropp2004greed,needell2010signal}. Among those,
particular attention has been paid to the orthogonal matching pursuit (OMP) algorithm 
\cite{tropp2007signal,pati1993orthogonal} which has an intuitive geometric interpretation and is characterized 
by high speed and competitive performance; numerous modifications of OMP that explore the trade-off 
between accuracy and speed have been proposed in literature 
\cite{donoho2012sparse,needell2009cosamp}. A related
Orthogonal Least-Squares (OLS) method \cite{chen1989orthogonal}, proposed as an identification 
algorithm for parameter estimation of generally multivariable non-linear systems which are linear in 
parameters, has recently been employed in compressed sensing \cite{soussen2013joint}. In general, OLS 
outperforms OMS in settings where the columns of $\A$ are non-orthogonal but it does so at a moderate
increase in complexity. The existing analysis and performance guarantees for OLS are limited to the
case of non-random measurements \cite{soussen2013joint,herzet2013exact,herzet2016relaxed}.

In this paper, we provide a result establishing that in the noiseless scenario where the coefficient matrix 
is drawn at random from a Gaussian or a Bernoulli distribution, with ${\cal O}\left(k\log(m)\right)$ linear random 
measurements OLS guarantees recovery of $\x$ with high probability. This result is comparable to those previously 
provided for OMP \cite{tropp2007signal,rangan2009orthogonal}. Moreover, we propose a generalization of 
OLS, the Generalized Orthogonal Least-Squares (GOLS), an efficient algorithm which relies on a recursive 
relation between the components of the optimal solution to (\ref{eq:1}) to select a pre-determined number of 
columns and provide performance superior to existing methods.


\section{Performance Guarantee for Orthogonal Least-Squares}

The OLS algorithm sequentially projects columns of $\A$ onto a residual vector and each time selects the 
column that leads to the smallest residual norm. Specifically, OLS chooses a new index $j_s$ as
\begin{equation*}
{j}_{s}=\argmin_{j \in {\cal I}}{\norm{\y - \A_{{\cal S}_{i-1}\cup\{j\}}\A_{{\cal S}_{i-1}\cup\{j\}}^{\dagger}\y}_2},
\end{equation*}
where $\cal I$ is the set of indices that are not yet selected. This procedure is computationally more expensive 
than OMP since in addition to solving the least-square problem to update the residual vector, orthogonal 
projection of each column needs to be found at each step of OLS. Perhaps in part due to this increase in 
complexity, OLS has not played as prominent role in sparse signal recovery literature as OMP did.

Note that the performances of OLS and OMP are identical when the columns of $\A$ are 
orthogonal.\footnote{In fact, orthogonality of the columns of $\A$ leads to a modular objective function in
(\ref{eq:2}), implying optimality of both methods.} It is beneficial to further clarify the difference between 
OMP and OLS. In each iteration of OMP, an element that best correlates with the current residual is chosen. 
OLS, on the other hand, selects a column that has the largest portion which is inexpressible by previously 
selected columns which, in turn, minimizes related approximation error. 

The following theorem states that for Gaussian and Bernoulli matrices with normalized columns, which are
often considered in compressed sensing problems, in the noiseless scenario OLS is with high probability 
capable of the exact recovery of sparse signals if the number of measurements grows linearly with the 
sparsity level and logarithmically with the dimension of the unknown signal.
\begin{theorem}
\textit{Suppose that $\x \in \R^m$ is an arbitrary sparse vector with sparsity level $k$. Consider a random 
matrix $\A \in \R^{n\times m}$ such that its entries are drawn uniformly and independently from either 
${\cal N}(0,1/n)$ or $\{+1\big\slash\sqrt{n},-1\big\slash\sqrt{n}\}$. Given the noiseless observation 
$\y=\A\x$, the OLS algorithm can recover $\x$ in $k$ iterations with probability of success exceeding 
$1-\delta$ if $n={\cal O}\left(k\log(m/\delta)\right)$ for some $0<\delta<\gamma$, where $\gamma$ is a 
positive constant which is independent of $n$, $m$, and $k$.}
\end{theorem}

The proof, which exploits the fact that the columns of A are spherically symmetric random vectors and
relies on Johnson-Lindenstrauss lemma, is omitted for brevity.


\section{Generalized Orthogonal Least-Squares}

To formulate generalized OLS, we start by establishing a recursive relation between the components of the 
optimal solution to the 
$l_0$-constrained least-squares problem. Let ${\bf B}_i$ denote the sub-matrix of $\bf A$ constructed 
by selecting $i$ of its columns and let ${\P}_{i}={\bf B}_{i}{\bf B}^{\dagger}_{i}$ denote the projection 
matrix onto the span of the columns of $\B_i$, where 
${\bf B}^{\dagger}_{i} = \left({\bf B}^{T}_{i}{\bf B}_{i}\right)^{-1}{\bf B}^{T}_{i}$  is the Moore-Penrose 
pseudo-inverse of $\B_i$. Then, after appending $\B_i$ with another column vector $\a$ to form
$\B_{i+1} = [\B_i \;\;\; \a]$, we can write 
\begin{equation} \label{eq:8}
\begin{aligned}
{\P}_{i+1} & = \B_{i+1} \left({\bf B}^{T}_{i+1}{\bf B}_{i+1}\right)^{-1}{\bf B}^{T}_{i+1} \\
&=
\begin{bmatrix}
    {\B}_i & \a 
\end{bmatrix}
\begin{bmatrix}
    {\B}^{T}_i{\B}_i   &  {\bf B}^{T}_i\a  \\
    \a^{T}{\B}_i       & \a^{T}\a
\end{bmatrix}^{-1}
\begin{bmatrix}
    {\B}^{T}_i        \\
    \a^{T}     
\end{bmatrix}\\
&\stackrel{(a)}{=}
\begin{bmatrix}
    {\B}_i & \P^\bot_i\a
\end{bmatrix}
\begin{bmatrix}
    \left({\bf B}^{T}_i{\B}_i\right)^{-1}      &  0  \\
    0      & \left(\a^{T}\P^\bot_i\a\right)^{-1}
\end{bmatrix}
\begin{bmatrix}
    {\B}^{T}_i        \\
    \a^{T}\P^\bot_i
\end{bmatrix}\\
&\stackrel{(b)}{=} \P_i+\frac{\P^\bot_i\a\a^T\P^\bot_i}{\norm{\P^\bot_i\a}^2_2}
\end{aligned}
\end{equation}
where ($a$) follows after the LDU decomposition of the intermediate matrix inverse, 
i.e., we use the identity \cite{kailath2000linear}
\begin{equation}
\begin{aligned}
\hspace{-0.4cm}\begin{bmatrix}
    {\bf A}       &  {\bf  E}  \\
    {\bf C}         & {\bf D}   
\end{bmatrix}^{-1} = 
\begin{bmatrix}
    {\bf I}       &  {\bf A}^{-1} {\bf E}   \\
    {\bf 0}      & {\bf I}   
\end{bmatrix} 
 \begin{bmatrix}
    {\bf A}^{-1}       &  {\bf 0}   \\
    {\bf 0}        & {\bf \Delta}^{-1} 
\end{bmatrix}
\begin{bmatrix}
    {\bf I}       &  {\bf 0}  \\
    {\bf C}{\bf A}^{-1}        & {\bf I}   
\end{bmatrix}
\end{aligned}
\nonumber
\end{equation}
where we identify ${\bf A}={\B}^{T}_i{\B}_i$, ${\bf E}={\bf B}^{T}_i\a$, ${\bf C}=\a^{T}{\B}_i $, 
${\bf D}=\a^{T}\a$, and ${\bf \Delta}={\bf D}-{\bf C}{\bf A}^{-1}{\bf E}$, and introduce 
${\P}_i^{\bot}=\I-{\P}_i$. The identity ($b$) follows from the idempotent property of the projection matrix.
Alternatively, we write (\ref{eq:8}) as
\begin{equation}
{\P}_{i+1}^{\bot}={\P}_i^{\bot}-\frac{{\P}_i^{\bot}\a\a^T {\P}_i^{\bot}}{\norm{{\P}_i^{\bot}\a}_2^2}.
\label{eq:perp}
\end{equation}
Note that, (\ref{eq:perp}) is related to order-recursive least-squares \cite{kay2013fundamentals}. However, this specific derivation makes it suitable for iterative sparse reconstruction applications.

Now, the OLS algorithm in each step selects a column with index $j_s$ from the set ${\cal I}$ of the 
previously non-selected columns according to
\begin{equation}
\begin{aligned}
{j}_{s}&=\argmin_{j \in {\cal I}}{\norm{\y - \A_{{\cal S}_{i-1}\cup\{j\}}\A_{{\cal S}_{i-1}\cup\{j\}}^{\dagger}\y}_2}\\
&\stackrel{(a)}{=}\argmin_{j \in {\cal I}}{\norm{\left(\I-\P_i\right)\y}_2^2}\\
&=\argmin_{j \in {\cal I}}{\y^T\y-\y^T\P_i\y-\y^T\P_i^T\y+\y^T\P_i^T\P_i\y}\\
&\stackrel{(b)}{=}\argmin_{j \in {\cal I}}{\y^T\y-\y^T\P_i\y}\\
&\stackrel{(c)}{=}\argmax_{j \in {\cal I}}{\y^T\P_{i-1}\y+\y^T\frac{\P^\bot_{i-1}\a_j\a^T_j\P^\bot_{i-1}}{\norm{\P^\bot_{i-1}\a_j}^2_2}\y}\\
&\stackrel{(d)}{=} \argmax_{j \in {\cal I}}{\frac{\norm{\y^T\P^\bot_{i-1}\a_j}^2_2}{\norm{\P^\bot_{i-1}\a_j}^2_2}}
=\argmax_{j \in {\cal I}}{\left|\y^T\frac{{\bf P}_{i-1}^{\bot}{\bf a}_{j}}{\norm{{\bf P}_{i-1}^{\bot}{\bf a}_{j}}_2}\right|}
\end{aligned}
\end{equation}
where ($a$) follows from the definition of $\P_i$, ($b$) is due to $P_i^\perp$ being an idempotent projection 
matrix, ($c$) follows from eq. (\ref{eq:8}), and ($d$) is due to the fact that $\y^T\P_{i-1}\y$ is not a function 
of the optimization variable. 


We propose a straightforward extension of OLS which selects multiple (say, $L$) columns of $\A$ 
in each step rather than choosing a single column, ultimately replacing the underdetermined 
$n \times m$ system of equations by an overdetermined $Lk \times k$ one. This strategy is 
motivated by the observation that the candidate columns whose projection onto the space orthogonal
to that spanned by the previously selected columns is strongly correlated with the observation
vector but not chosen in the current step of OLS will likely be selected in subsequent steps of the 
algorithm; therefore, selecting several ``good" candidates in each step accelerates the selection 
procedure and enables sparse reconstruction with fewer steps (and, therefore, fewer calculations
of the mutual correlations needed to perform the selection). More specifically, the proposed 
generalized OLS algorithm performs the following: in each step, the algorithm select $L$ columns 
of matrix $\A$ such that their normalized projection onto the orthogonal complement of the 
subspace spanned by previously chosen columns have the highest correlation with the observation 
vector among the non-selected columns. After such columns are identified, we update the orthogonal 
projection matrix by repeatedly applying (\ref{eq:perp}) $L$ times. We continue until a stopping 
criterion is met. Generalized orthogonal least-squares algorithm is formalized as Algorithm 1.

\subsection{Computational complexity} 

To analyze the computational complexity of GOLS, note that its first step involves a matrix-vector 
multiplication and a vector inner product; the computational cost of these two is dominated by the 
former and thus requires ${\cal O}\left(mn^2\right)$ operations. The other operations are those in 
Step 3 where a matrix-vector multiplication and a matrix addition, needing ${\cal O}\left(n^2\right)$ 
calculation, need to be repeated $\min\{k,\lfloor \frac{n}{L}\rfloor\}$ times. Therefore, the aggregate 
cost of operations in this step is ${\cal O}\left(mkn^2\right)$. Finally, finding the estimate $\hat{\x}$ 
entails solving a least-squares problem which can be implemented with a small cost 
${\cal O}\left(kn\right)$ by relying on a QR factorization of $\A_{{\cal S}_{k}}$. 
Therefore, the total complexity of the algorithm is ${\cal O}\left(mkn^2+kn\right)$.

\begin{figure*}[t]
\begin{subfigure}[b]{0.32\textwidth}
  \centering
    \includegraphics[width=1\textwidth]{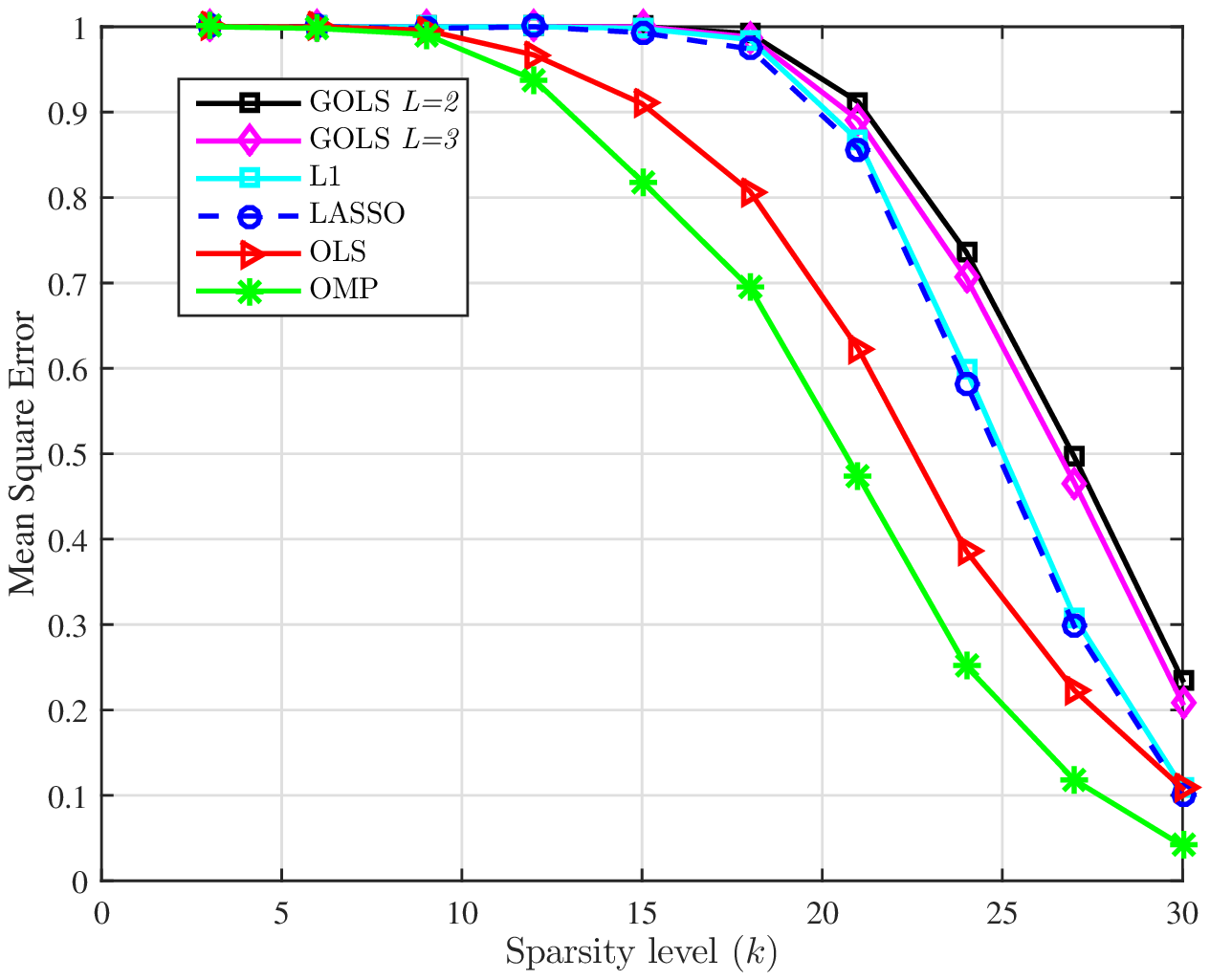}\quad\caption{ERR}
    \end{subfigure}
    \begin{subfigure}[b]{.32\textwidth}
  \centering
    \includegraphics[width=1\textwidth]{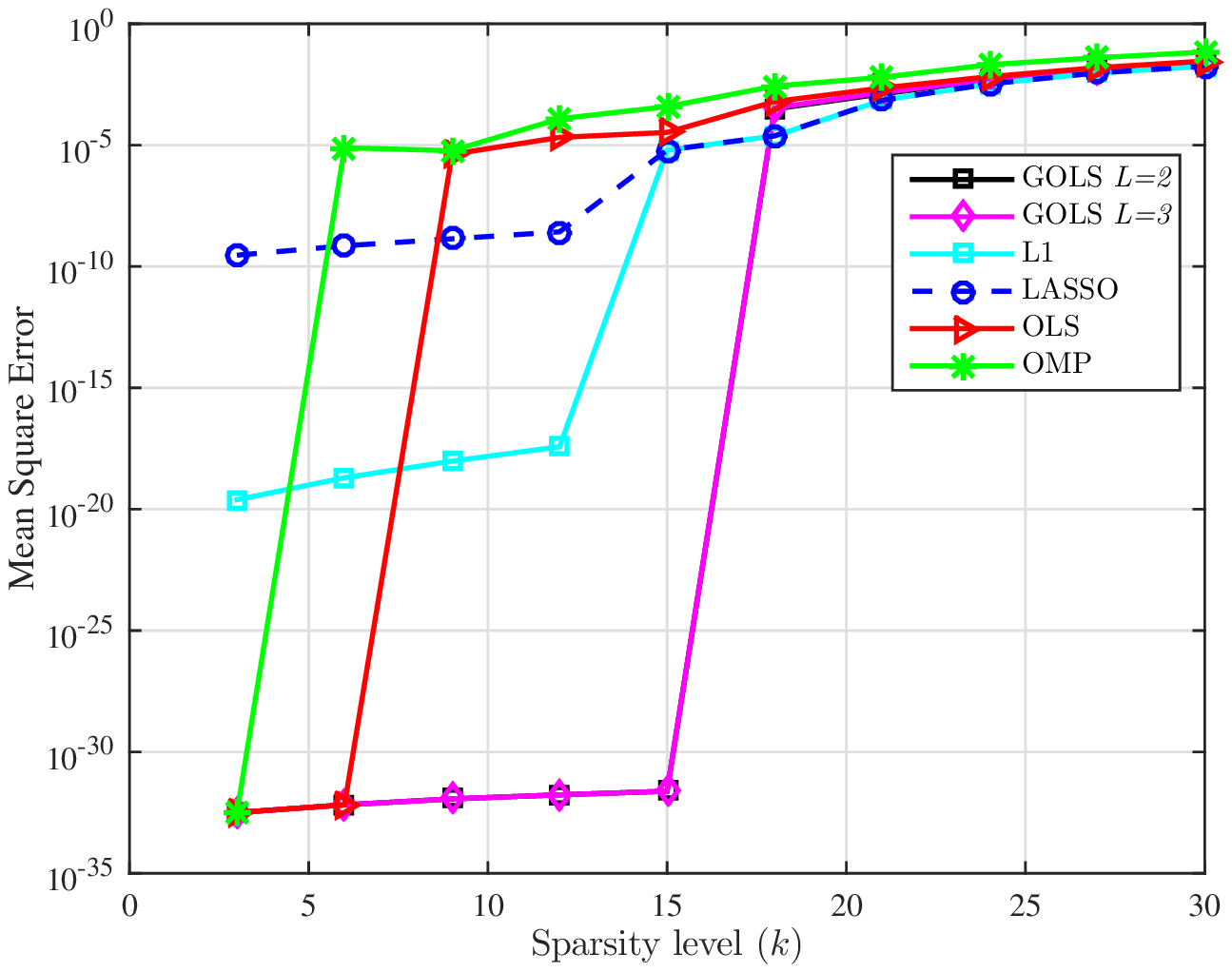}\quad\caption{MSE}
    \end{subfigure}
    \begin{subfigure}[b]{.32\textwidth}
  \centering
    \includegraphics[width=1\textwidth]{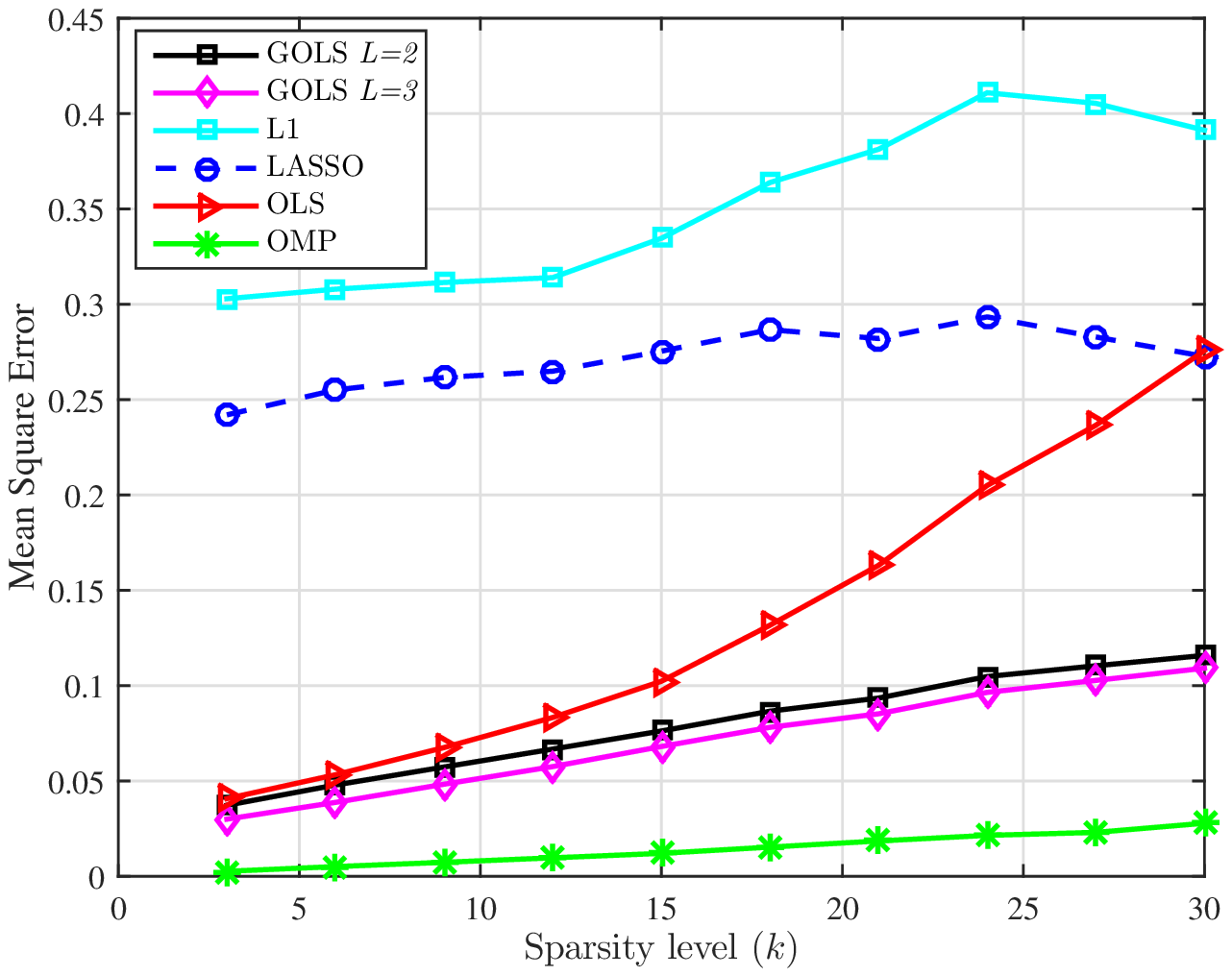}\quad\caption{Running time}
\end{subfigure}
\caption{Performance comparison of GOLS, OLS, OMP, $l_1$-norm minimization and LASSO
for $n=64$, $m=128$, $\A$ having Gaussian ${\cal N}(0,1/n)$ entries, and the $k$ non-zero
components of $\x$ drawn from ${\cal N}(0,1)$ distribution.}
\end{figure*}
\begin{figure*}[t]
\begin{subfigure}{.32\textwidth}
  \centering
    \includegraphics[width=1\textwidth]{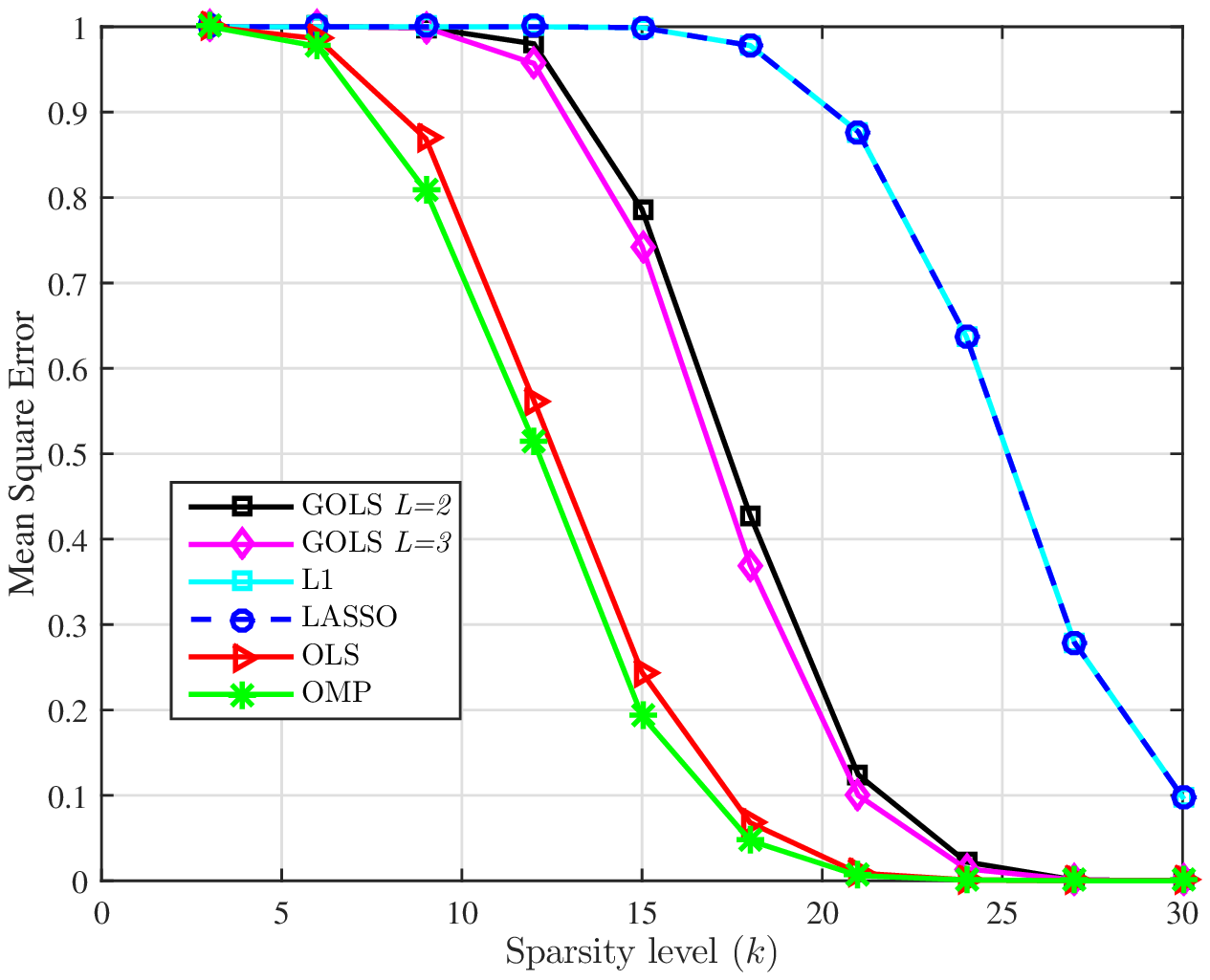}\quad\caption{ERR}
    \end{subfigure}
    \begin{subfigure}{.32\textwidth}
  \centering
    \includegraphics[width=1\textwidth]{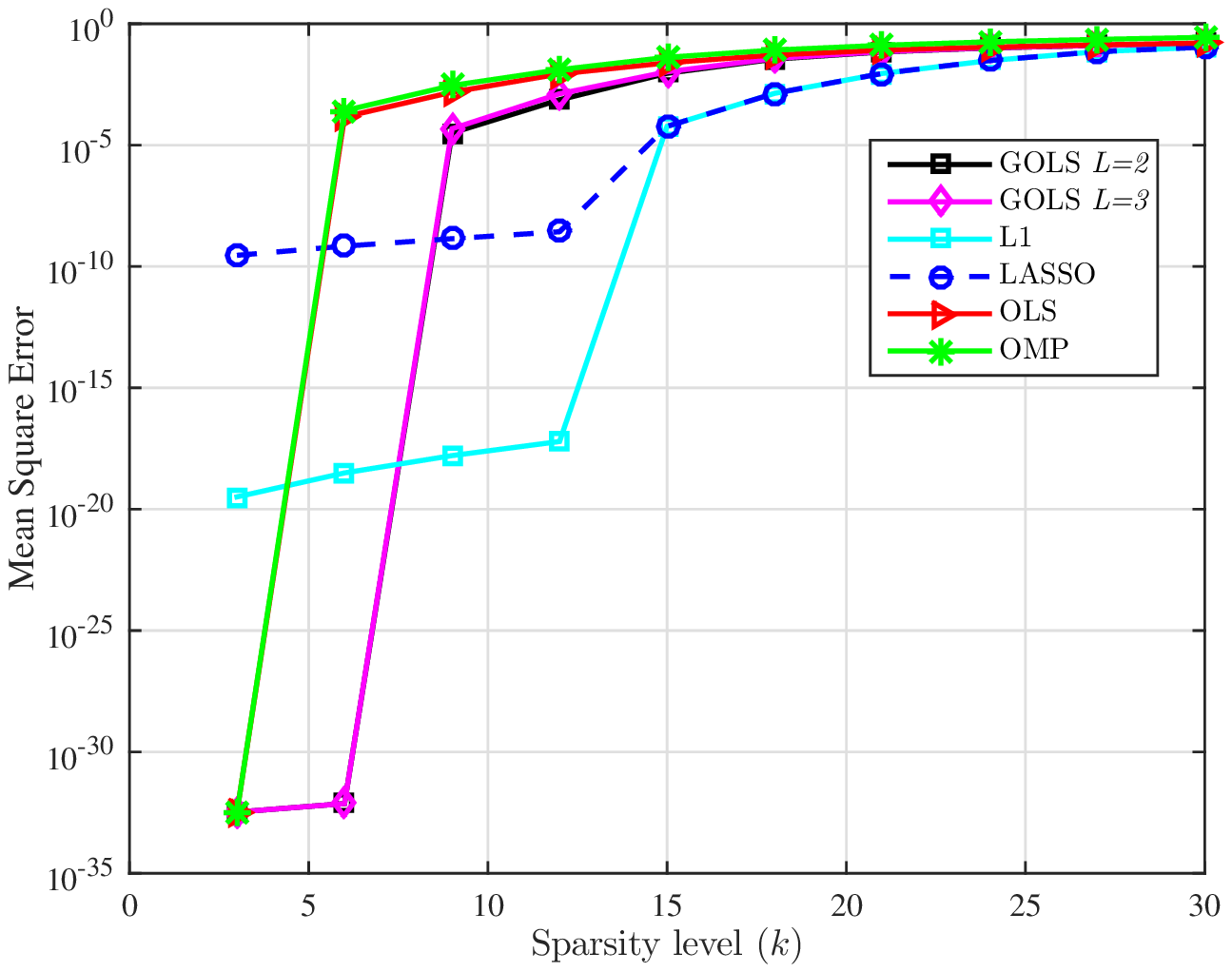}\quad\caption{MSE}
        \end{subfigure}
        \begin{subfigure}{.32\textwidth}
  \centering
    \includegraphics[width=1\textwidth]{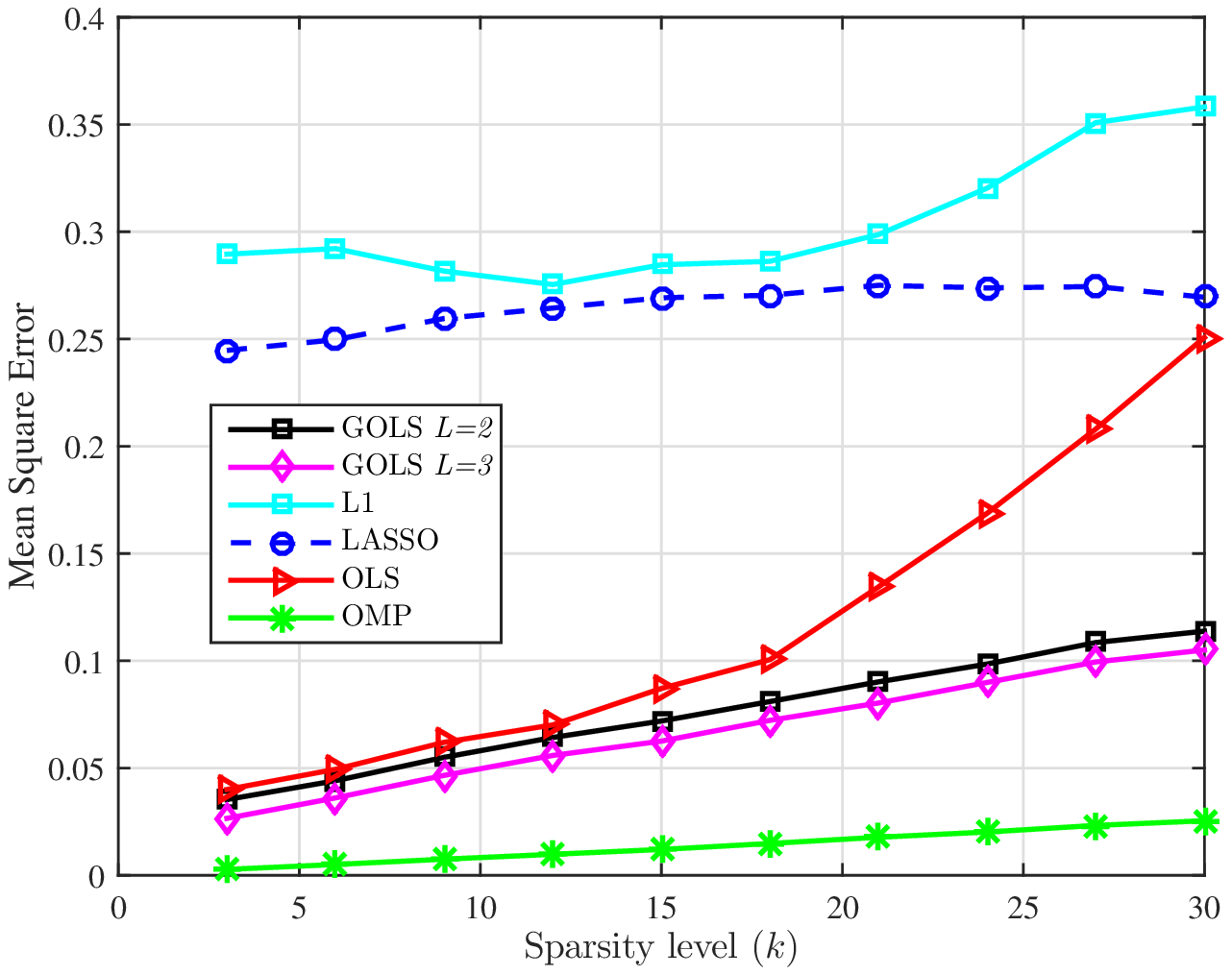}\quad\caption{Running time}
\end{subfigure}
\caption{Performance comparison of GOLS, OLS, OMP, $l_1$-norm minimization and LASSO
for $n=64$, $m=128$, $\A$ having Gaussian ${\cal N}(0,1/n)$ entries, and the $k$ non-zero
components of $\x$ randomly and equally likely set to $1$ or $-1$.}
\end{figure*}
\begin{figure*}[t]
\begin{subfigure}{0.32\textwidth}
  \centering
    \includegraphics[width=1\textwidth]{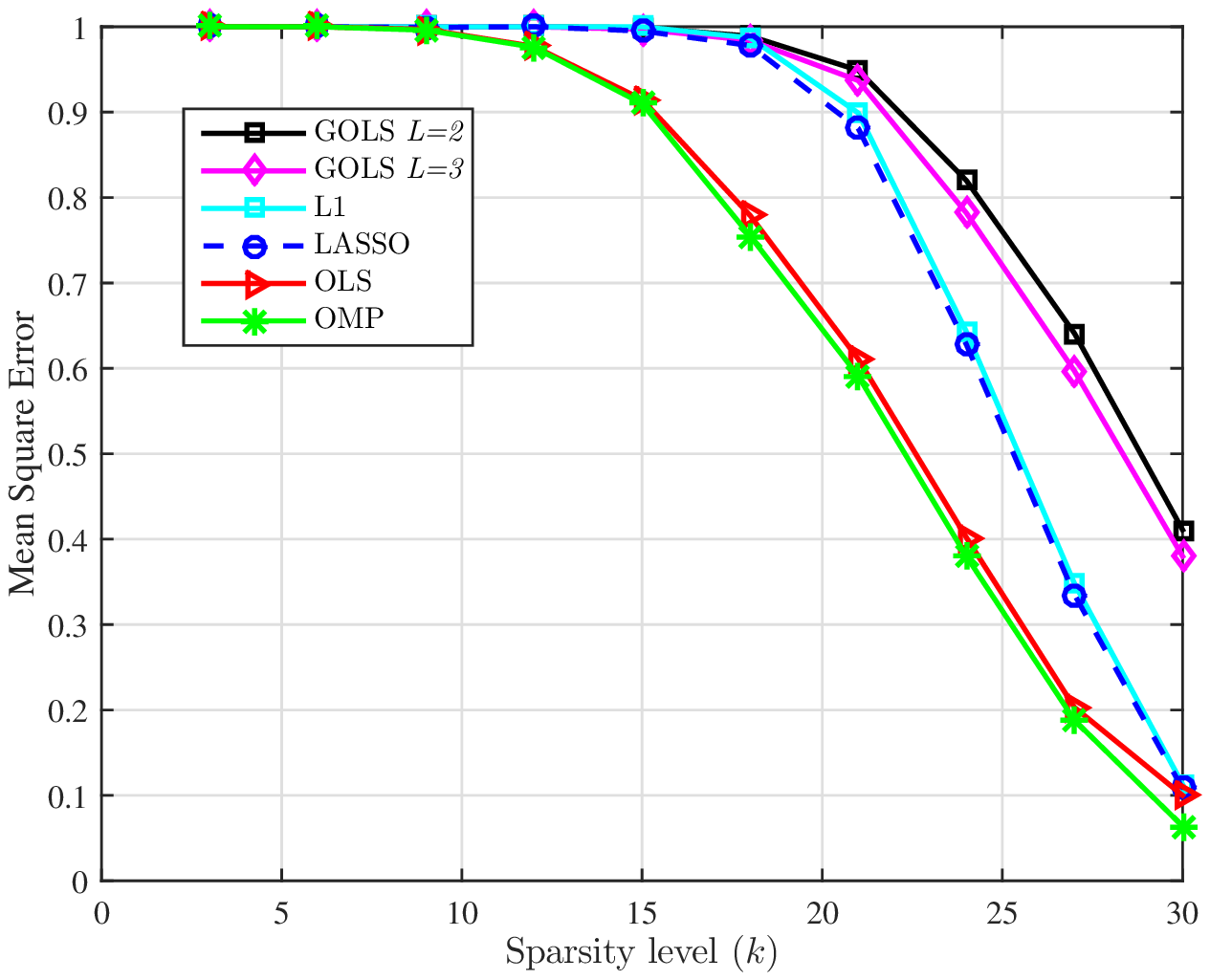}\quad\caption{ERR}
    \end{subfigure}
    \begin{subfigure}{.32\textwidth}
  \centering
    \includegraphics[width=1\textwidth]{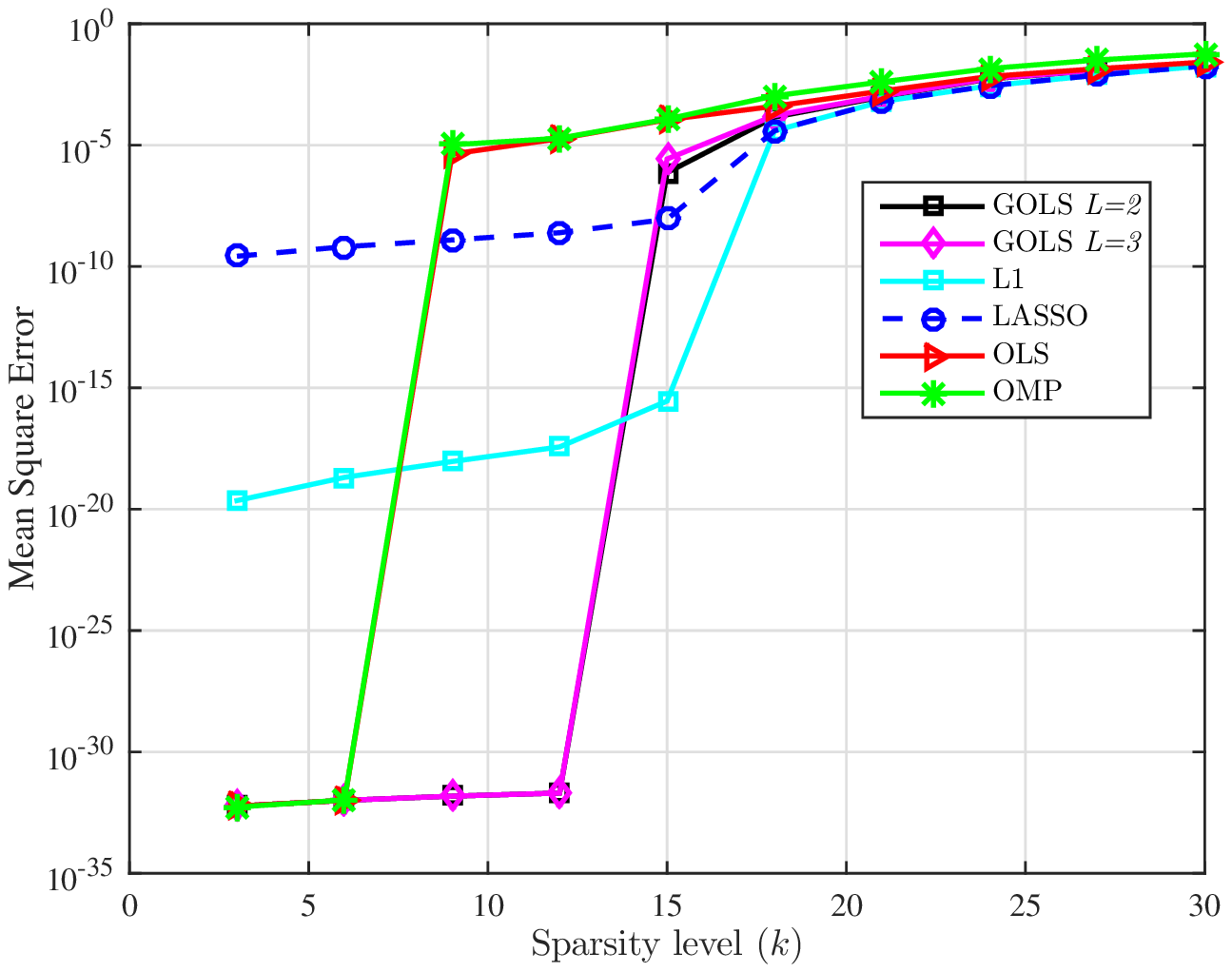}\quad\caption{MSE}
        \end{subfigure}
        \begin{subfigure}{.32\textwidth}
  \centering
    \includegraphics[width=1\textwidth]{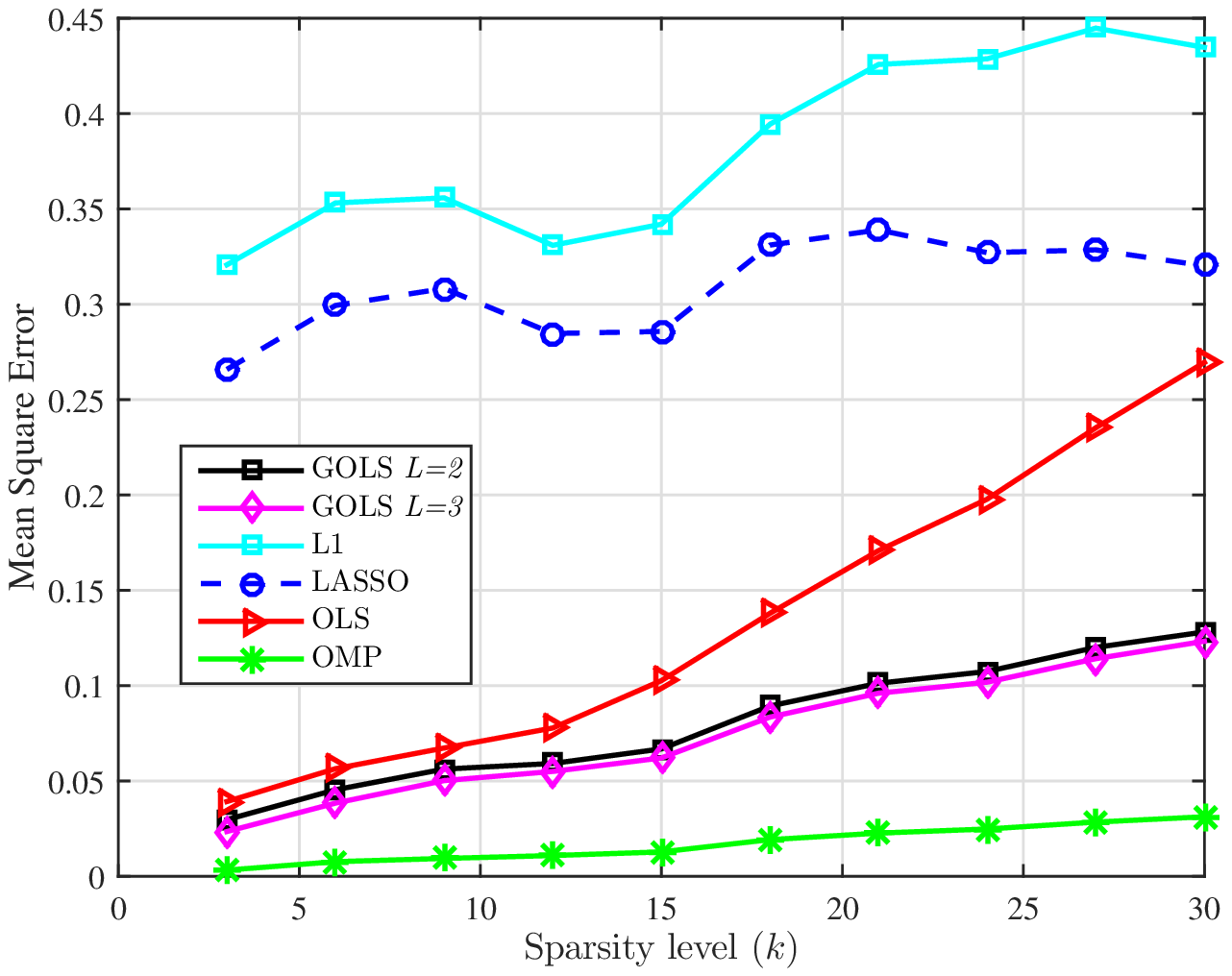}\quad\caption{Running time}
\end{subfigure}
\caption{Performance comparison of GOLS, OLS, OMP, $l_1$-norm minimization and LASSO
for $n=64$, $m=128$, $\A$ having uniformly i.i.d entries from $\{+1\big\slash\sqrt{n},-1\big\slash\sqrt{n}\}$, and the $k$ non-zero
components of $\x$ drawn from ${\cal N}(0,1)$ distribution.}
\end{figure*}

\begin{algorithm}
\vspace{0.1cm}
\begin{tabularx}{\textwidth}{l>{$}c<{$}X}
\textbf{Input:}  \hspace{0.85cm}observation $\y$, coefficient matrix ${\bf A}$, sparsity $k$ 
\vspace{0.1cm}\\
\textbf{Output:} \hspace{0.5cm} recovered support ${\cal S}_k$, estimated signal $\hat{\x}_{k}$\vspace{0.1cm}\\
\textbf{Initialize:} \hspace{0.3cm}  ${\cal S}_0=\oldemptyset$, ${\bf P}_0^{\bot}={\bf I}$, ${\cal I} = \{1,2,\dots,m\}$
\end{tabularx}
\begin{algorithmic}
\FOR  { $i=1$ to $\min\{k,\lfloor \frac{n}{L}\rfloor\}$  }\vspace{0.1cm}
\STATE 1. Select $\{i_{s_1},\dots,i_{s_L}\}$ corresponding to $L$ largest terms:
$\left|\y^T\frac{{\bf P}_{i-1}^{\bot}{\bf a}_{j}}{\norm{{\bf P}_{i-1}^{\bot}{\bf a}_{j}}_2}\right|$ for $j \in {\cal I}$\vspace{0.1cm}\\
2. ${\cal S}_{i}={\cal S}_{i-1}\cup\{i_{s_1},\dots,i_{s_L}\}$ , ${\cal I} = {\cal I} \backslash {\cal S}_{i}$\vspace{0.05cm}\vspace{0.1cm}\\
3. ${\bf D}={\bf P^{\bot}}_{i-1}$ \vspace{0.1cm}\\
\FOR { $l=1$ to $L$ }\vspace{0.1cm}
\STATE  ${\bf d}=\frac{{\bf D}{\bf a}_{j_{s_l}}}{\norm{{\bf D}{\bf a}_{j_{s_l}}}_2}, {\bf D}={\bf D}-{\bf d}{\bf d}^T$\vspace{0.1cm}
\ENDFOR\vspace{0.1cm}\\
${\bf P}_{i}^{\bot}={\bf D}$ \vspace{0.1cm}\\
\ENDFOR \vspace{0.1cm}
\STATE $\hat{\x}_k=\A_{{\cal S}_{k}}^{\dagger}\y$
\end{algorithmic}
\caption{Generalized Orthogonal Least-Squares}
\label{algo:2}
\end{algorithm}
\section{SIMULATION RESULTS} \label{sec:sim}
To evaluate the algorithm, we compared its performance with four other sparse recovery algorithms as 
a function of the sparsity level $k$. In particular, we considered OMP, OLS, $l_1$-norm minimization 
\cite{candes2006robust}, and Least Absolute Shrinkage and Selection Operator (LASSO) 
\cite{tibshirani1996regression}. As typically done in benchmarking tests \cite{dai2009subspace}, we used 
CVX \cite{grant2008cvx} to implement the $l_1$ minimization and LASSO. The tuning parameter in LASSO 
is found by means of 10-fold cross validation. We draw entries of the coefficient matrix $\A$ from two
distributions. First, we generate entries of $\A$ by drawing independently from a Gaussian distribution with 
zero-mean and variance $n^{-1}$. We then consider two different scenarios for this choice of coefficient 
matrix: (1) the non-zero elements of $\x$ are independent and identically distributed normal random variables, 
and (2) the non-zero components of $\x$ are drawn uniformly from alphabet $\{+1,-1\}$. Second, the entries 
of $\A$ are drawn independently and uniformly from $\{+1\big\slash\sqrt{n},-1\big\slash\sqrt{n}\}$. In all 
settings, the locations of non-zero entries of $\x$ are drawn uniformly at random. The number of equations 
is $n=64$, the dimension of $\x$ is $m=128$; the experiment is repeated 1000 times. Performance of each 
algorithm is characterized by three metrics: (i) exact recovery rate (ERR), defined as the fraction of the 
correctly recovered signal components, (ii) mean-square error (MSE), measuring the distance between the 
unknown signal and its estimate, and (iii) the running time of the algorithm. Results for the Gaussian 
coefficient matrices are illustrated in Fig. 1 and Fig. 2. Fig. 1 shows the performance of the algorithms for 
non-zero values of $\x$ being normally distributed while Fig. 2 corresponds to the second scenario. Fig. 3 
shows the performance of the methods for $\A$ being constructed according to the second option while 
the non-zero values of $\x$ are normally distributed. As can be seen from Fig. 1 and Fig. 3, the generalized 
OLS (GOLS) outperforms all the competing methods in terms of the exact recovery rate, and is better than 
OLS and OMP in terms of the MSE. Moreover, the runtimes of GOLS is 2nd only to OMP but the accuracy 
of the latter is significantly worse than that of GOLS. In the case of $\{+1,-1\}$ non-zero entries of $\x$ 
studied in Fig. 2, $l_1$-norm/LASSO methods perform the best (and are the slowest) while the GOLS 
offers reasonably accurate performance at relatively high speed.



\section{CONCLUSION} \label{sec:concl}
We show that for Gaussian and Bernoulli coefficient matrices, Orthogonal Least-Squares (OLS) is with 
high probability guaranteed to recover any sparse signal from a low number of random linear measurements. 
Moreover, we introduced a greedy algorithm for sparse linear regression that generalizes OLS and forms 
the subset of features (i.e., columns of a coefficient matrix in an underdetermined system of equations) by 
sequentially selecting multiple candidate columns. Since multiple indices are selected without additional cost, 
the running time of the algorithm is reduced compared to OLS. Thus, generalized OLS is more favorable than 
convex optimization based methods whose complexity grows faster with the dimension of the problem, i.e., 
$n$ and $m$. Simulation studies demonstrate that the generalized orthogonal least-squares algorithm 
outperforms competing greedy methods, OLS and OMP, while being computationally more efficient than 
$l_1$-norm minimization and LASSO. 
\bibliographystyle{IEEEbib}\small
\bibliography{IEEEabrv,refs}

\end{document}